\documentclass[conference]{IEEEtran}

\usepackage{graphicx}
\usepackage{booktabs}
\usepackage[font=small,skip=0pt]{caption}
\usepackage{lipsum}
\usepackage{stfloats}
\usepackage{algorithm}
\usepackage{algorithmic}
\usepackage{amsmath}
\usepackage{commath}
\usepackage{caption}
\usepackage[bookmarks=false]{hyperref}
\usepackage{array}
\usepackage{subfig}
\usepackage{gensymb}
\usepackage{makecell}
\usepackage{authblk}
\usepackage{pgfplots}
\usepackage{tikz}
\usepackage{xcolor}
\usepackage{pgfplotstable}

\usetikzlibrary{patterns}
\pgfplotsset{compat=newest}

\usetikzlibrary{decorations.pathmorphing}
\usetikzlibrary{arrows,positioning,shapes}
\usepgfplotslibrary{groupplots}
\pgfplotsset{compat=1.5}
\newcolumntype{C}[1]{>{\centering\arraybackslash}p{#1}}
\newcolumntype{L}[1]{>{\raggedright\arraybackslash}m{#1}}

\makeatletter
\renewcommand\AB@affilsepx{, \protect\Affilfont}
\makeatother

\graphicspath{ {Figures/} }
\definecolor{Blue1} {RGB}{128, 177, 211}
\definecolor{Green1}{RGB}{179, 222, 105}
\definecolor{Red1}  {RGB}{251, 128, 114}
\definecolor{Blue2} {RGB}{0, 51, 102}
\definecolor{Green2}{RGB}{150, 190, 105}
\definecolor{Green3}{RGB}{0, 100, 0}
\definecolor{Red2}  {RGB}{240, 50, 50}
\definecolor{BLACK}  {RGB}{0, 0, 0}

\begin{document}


\IEEEoverridecommandlockouts
\IEEEpubid{\begin{minipage}[t]{\textwidth}\ \\[2pt]
        \normalsize{978-1-7281-2437-7/19/\$31.00 \copyright 2019 IEEE}
\end{minipage}}

\title{\fontsize{23}{23}\selectfont Comparing Energy Efficiency of CPU, GPU and FPGA Implementations for Vision Kernels}
\author[*]{Murad Qasaimeh}
\author[$\dag$]{Kristof Denolf}
\author[$\dag$]{Jack Lo}
\author[$\dag$]{Kees Vissers}
\author[*]{Joseph Zambreno}
\author[*]{Phillip H. Jones\vspace{-1em}}

\affil[*]{Iowa State University, IA, USA}
\affil[$\dag$]{Xilinx Research Labs, CA, USA\vspace{-1em}}

\maketitle

\begin{abstract}

Developing high performance embedded vision applications requires balancing run-time performance with energy constraints.  Given the mix of hardware accelerators that exist for embedded computer vision (e.g. multi-core CPUs, GPUs, and FPGAs), and their associated vendor optimized vision libraries, it becomes a challenge for developers to navigate this fragmented solution space. To aid with determining which embedded platform is most suitable for their application, we conduct a comprehensive benchmark of the run-time performance and energy efficiency of a wide range of vision kernels. We discuss rationales for why a given underlying hardware architecture innately performs well or poorly based on the characteristics of a range of vision kernel categories.  Specifically, our study is performed for three commonly used HW accelerators for embedded vision applications: ARM57 CPU, Jetson TX2 GPU and ZCU102 FPGA, using their vendor optimized vision libraries: OpenCV, VisionWorks and xfOpenCV.  Our results show that the GPU achieves an energy/frame reduction ratio of 1.1--3.2$\times$ compared to the others for simple kernels.  While for more complicated kernels and complete vision pipelines, the FPGA outperforms the others with energy/frame reduction ratios of 1.2--22.3$\times$.  It is also observed that the FPGA performs increasingly better as a vision application's pipeline complexity grows.\vspace{1mm}


\end{abstract}

\begin{IEEEkeywords}
Embedded Vision, GPUs, FPGAs, CPUs, Energy Efficiency. \vspace{-1em}
\end{IEEEkeywords}

\section{Introduction}

Image sensors are increasingly becoming an essential component of a wide range of embedded system applications, such as: smartphones, autonomous cars, drones. This trend is a driving force for the development of energy-efficient image processing solutions.  Energy-efficient image processing is especially important for tightly energy-constrained real-time embedded systems, as often their limited communication power budget or communication capabilities preclude them from streaming images to more powerful computing entities.

Both industry and academia have explored the development of acceleration engines to help meet the needs of embedded vision applications.  Three common types of such accelerators are benchmarked in this case study: multicore CPUs, Graphic Processing Units (GPUs), and Field Programmable Gate Arrays (FPGAs).   Each of these accelerators take a different approach to accelerating embedded vision applications.  Multi-core CPUs make use of  SIMD instruction extensions, such as: the ARM NEON SIMD engine, Intel's family of SSE, and  dedicated vision processing units (VPU), such as Myriad \cite{ionica2015movidius}.  The multi-threading programming model has made GPUs highly popular in this domain.  GPUs provide massively parallel execution resources and high memory bandwidth. However, their high performance comes at the cost of high power dissipation \cite{collange2009power}.  FPGAs offer opportunities for exploiting low-level fine-grained parallelism by customizing data paths to the requirements of a specific algorithm/application \cite{giefers2016analyzing}.

Embedded vision applications can exhibit vastly different performance characteristics depending on their underlying hardware accelerator platform \cite{che2008accelerating}.  This varying behavior fundamentally stems from differences in accelerator micro-architectures, middleware support, and programming styles. This mixture of factors makes choosing the best application-to-accelerator mapping a nontrivial task for embedded vision application developers. They must take into consideration metrics, such as expected runtime performance, energy-efficiency, and programmability.  An additional challenge facing developers is partitioning vision pipelines into phases that can run on available accelerators in the most efficient and cost-effective manner.
In order to clearly understand how different hardware architectures may impact the performance of vision kernels, we analyze the performance of such accelerators for different vision kernels. In this paper, we evaluate the performance of three popular HW accelerators for vision applications: the ARM57 CPU, Jetson TX2 GPU, and ZCU102 FPGA. We propose and evaluate an easily reproducible benchmarking approach that only uses publicly available vision libraries: OpenCV, Nvidia VisionWorks and xfOpenCV, without adding any special platform specific code. All benchmark code is available at: https://github.com/isu-rcl/cvBench.

\textit{Contributions}. In this work, we benchmark the performance of standard vision kernels on low-power embedded platforms. The main contributions of this paper are: (1) Benchmark representative vision kernels and complete pipelines for the ARM57 CPU, Nvidia Jetson TX2 (GPU-accelerated) and Xilinx UltraScale (FPGA-accelerated), (2) Insight into the reasons behind the observed run-time, power, and energy consumption performance for each evaluated platform, and (3) An energy efficiency comparison between the three hardware accelerator platforms evaluated in terms of energy delay product (EDP).

\textit{Organization}.
The remainder of this paper is organized as follows. Section \ref{sectionRelatedWork} reviews related work. Section \ref{sectionBackground} presents six categories of vision algorithms, and provides insights into the architectural differences between the hardware accelerators evaluated. In Section \ref{sectionMethodology}, we present the performance metrics used in this study and provide a detailed description of our measurement methodology. In Section \ref{sectionResults}, we discuss our experimental results and observations. Finally, Section \ref{sectionConclusions} concludes the paper with outlooks for future work.

\section{Related Work} \vspace{1mm} \label{sectionRelatedWork}

Most prior studies focus solely on comparing the performance of single vision kernels on embedded GPUs and FPGAs, with a few exceptions as discussed below. 
The comparison study in \cite{cong2018understanding} analyzed the performance efficiency of FPGAs and GPUs on the GPU-friendly benchmark suite (Rodinia). They ported 15 of its kernels using Vivado HLS for the FPGA and OpenCL for host programs. The platforms used were a Virtex-7 FPGA  and Tesla K40c GPU. Although this study includes some vision kernels such as: GICOV, Dilate, SRAD and MGVF, it was not mainly focused on benchmarking vision algorithms; it included other kernels for data mining, fluid dynamic, and physics simulation, etc \cite{che2009rodinia}.

Other comparison studies focused on a subset of vision kernels. For example, the study in \cite{cooke2015tradeoff} and \cite{fowers2012performance} evaluated the performance of sliding window applications on FPGAs, GPUs and multi-core CPUs. They compared the performance of three applications: Sum of Absolute Differences (SAD), 2D convolution, and correntropy. The platforms used in their study were an Altera Stratix IV FPGA, an NVIDIA GeForce GTX 560, and an Intel Xeon Core i7.
Another study in \cite{brugger2015quantitative} focused on comparing the performance of morphological image filtering operations. The authors utilized the OpenCV library for a CPU and GPU (cv::CUDA module). For the FPGA platform, they used Vivado HLS video libraries and hand-optimized implementations. The platforms used in their study were the Zynq 7020 FPGA, Tegra K1, and Intel core i7. 
The work in \cite{fykse2013performance} also focused only on applications such as normalized cross correlation and finite impulse response (FIR) filters.  The study's evaluation included development time, component cost, and power consumption.

In our work, we evaluate the run-time performance and energy efficiency of different embedded hardware solutions over a wide range of standard vision kernels. We provide rationale for when and why specific hardware platforms perform well or poorly for specific vision kernels.

\section{Background}  \label{sectionBackground}

In this section, we first present the characteristics of the hardware accelerators evaluated in this study. Then, we briefly discuss the three vision libraries widely used with these accelerators. Finally, we group vision kernels into categories based on their characteristics to understand the implications of the underlying hardware architectures on the performance of the kernels in their respective categories.

\subsection{Embedded Platforms}   

\textbf{1. Central Processing Unit (CPU):} Modern CPUs are able to preform SIMD (Single Instruction, Multiple Data) instructions using multiple ALUs.  These SIMD instruction sets are useful in the context of image processing, where operations are often repetitively applied to a continuous stream of data. This is particularly true in the context of computer vision, where most operations are performed over the entire image.  Examples of SIMD architectures are: ARM NEON SIMD engine and Intel's streaming SIMD extensions (SSE).  

\textbf{2. Graphic Processing Unit (GPU):} 

As compared to general purpose CPUs, which have developed SIMD instruction extensions to help parallelize image processing type tasks, GPUs have taken the direction of evolving into  a specialized SIMD architecture.  This specialization has led to GPUs having simpler processing cores than high-performance general purpose CPUs.  For example, they have simpler control logic, typically no branch prediction or prefetch, and small per-core memory.  Simpler computing cores allow GPUs to pack many more cores into a chip than a general purpose CPU.    GPU architectures perform extremely well on workloads that have little to no branching conditions or data dependences. Additionally, GPU architectures have specialized their memory architecture to support high-speed data streaming for image processing.  For example, the L2 cache in the Jetson TX2 (Pascal GPU) is 2048 KB, which can fit a 1080p grayscale image.

\textbf{3. Field Programmable Gate Array (FPGA):}
Instead of having a fixed processor-like design, FPGAs consist of an array of logic blocks, DSPs, on-chip BRAMs, I/O pads, and routing channels. 
In FPGA, custom data paths can be architected to stream pixels directly between computing units without needing to read/write from/to external memory. Moreover, the distributed on-chip BRAMs can be used to exploit data locality in vision kernels by keeping pixels on-chip (e.g Zynq UltraScale MPSoC FPGA has 32.1 Mb on-chip memory).
With FPGAs, developers need to ensure that their customized designs meet timing and space requirements.   

\subsection{Computer Vision Libraries}
A number of vision libraries have been optimized to target the hardware platforms discussed in the previous section. In this work, we focused on the most complete and commonly used libraries, as follows:    

\textbf{1. OpenCV:}
OpenCV is the de-facto standard C/C++ library for image and vision processing  \cite{itseez2018opencv}. It is used by the computer vision community to create desktop and embedded vision applications. OpenCV  has bindings for languages such as Python and Java. The latest version of OpenCV (at the time of writing this paper) is 4.0.  

\textbf{2. Nvidia VisionWorks:}
VisionWorks is a toolkit for computer vision and image processing released by Nvidia in 2015 \cite{brill2018nvidia}. It implements and extends the OpenVX standard, and is optimized for CUDA-capable GPUs. VisionWorks provides three programming models: immediate mode, graph mode and CUDA API. The latest version of VisionWorks is 1.6.   

\textbf{3. Xilinx xfOpenCV:}
The xfOpenCV library is a set of OpenCV functions optimized for Zynq and Zynq UltraScale devices  by Xilinx \cite{xfopencv2018}. It was first released in 2017, as part of the Xilinx reVISION stack. It has been implemented using HLS to work in their SDx development environment and provides a software interface for building vision pipelines on FPGAs. The latest version of the xfopenCV library is 2018.3.

\subsection{Categories of Vision Kernels} 
Computer vision algorithms can be grouped into six categories based on their functionality. The complexity of these kernels grows over the first five categories. The last category includes composite kernels, which are composed of kernels from the other categories. The following  discusses each category in more detail: 

\textbf{1. Input Processing:} The kernels in this group are usually used as pre-processing steps. They include  simple arithmetic operations to change the input format or number of channels into a desired format.  Some examples of these kernels are: channel combine, channel extract, color conversion, and bitdepth conversion.

\textbf{2. Image Arithmetic:} Image arithmetic applies standard arithmetic/logic operations to one or more images. Because of the multi-dimensional nature of these pixel based operations, these kernels can  benefit from highly parallel hardware architectures, such as GPUs and FPGAs. Furthermore, the data being processed is very localized; the algorithms can be  distributed among different processing units without concerns of data dependencies. These operations include: thresholding, absolute difference, addition/subtraction, bitwise and/or/xor/not, multiplication, accumulate, accumulate squared, and accumulate weighted. 

\textbf{3. Image Filters:} 
These algorithms compute the correlation between an input image and a kernel (small matrix of fixed-size). The data in these algorithms are local to the size of the kernel which is different from the arithmetic case where the operations were performed on a pixel basis. When the underlying hardware has enough local memory to accommodate the kernel size, the algorithm is still easily distributed among parallel  processing units. On the other hand, nonlinear filters are more irregular as they have branching conditions. This impedes their decomposition into parallel blocks. These kernels include: filter2D, box filter, erode, dilate, median, pyramid up, and pyramid down.
 
\textbf{4. Image Analysis:} Analytic kernels are typically used to understand  characteristics of an image, such as color distribution, mean, maximum and minimum pixel value, etc. Also, they are usually placed at the end of vision pipelines to reduce the image into a decision variable (min/max locations).
These kernels are filled with branching conditions and complex memory access patterns that negatively impact their performance on CPUs and GPUs. These operations include: histogram, mean/std, min/max location, table lookup, histogram equalization, and integral image.

\textbf{5. Geometric Transformation:} Transformations in  geometric space are essential to understanding the 3D world through the lens of a 2D image sensor.
These kernels include matrix multiplication that map effectively into highly parallel architectures composed of simple computing blocks (e.g. GPU). While these kernels are simple, their performance is negatively affected by irregular memory access patterns. These kernels include: remap, resize, affine warp, and perspective warp.    

\textbf{6. Composite Kernels:} 
The kernels in this category are composed in part of kernels from the previously described categories. Examples of these composite kernels are: feature extraction, stereo block matching, and optical flow. \textit{Feature extraction}  is used to find  interesting pixels in an image. Once features are extracted, they are no longer  stored as a continuous block of adjacent pixels in memory. This forces other kernels to load non-continuous memory addresses, which may hinder parallelism performance. \textit{Stereo block matching} uses two cameras, with known position and characteristics, to compute disparity by comparing overlapped regions, leading to a high computational load. \textit{Optical flow} is used to estimate the apparent motion of objects between two consecutive images. Optical flow can be  computed for each pixel (dense) or a subset of pixels (sparse).


\section{Experimental Methodology} \label{sectionMethodology}\vspace{-1mm}

This section describes the performance metrics and measurement techniques used, and introduces our study's benchmarking approach.

\subsection{Performance Metrics} \vspace{-1mm}

Selecting proper metrics is essential for assessing the energy efficiency of vision kernels running on different hardware accelerators. These metrics should provide  meaningful interpretation and a fair way for comparison. In this subsection, we discuss the evaluation metrics used in our study:

\textbf{Run-time:}
Run-time performance of vision kernels can be evaluated by measuring the elapsed time (delay time) between the start and end  of a kernel's code. We use a high resolution timer to accurately measure  elapsed time while executing on HW accelerators. In our study, we measured the execution time only, and excluded the time required for copying images from/to the external memory of CPUs and GPUs, and the time  for configuring datamovers in FPGAs.

\textbf{Energy:}
Energy consumption per frame quantifies the amount of electrical energy dissipated by hardware accelerators to perform a kernel's operations on one frame. It is measured as the power consumed during the delay time to process a frame.
Device power can be divided in two parts: (1) Static power: represents the amount of power consumed when no active computation is taking place (system is idle), (2) Dynamic power: represents the amount of power consumed above the static power level when the system is computing.

\textbf{Energy-Delay Product (EDP):}
Run-time or energy per frame alone do not show the entire picture. A hardware platform can be extremely low power while being too slow to be of  practical use. The Energy Delay Product (EDP) \cite{horowitz2005scaling} metric takes into account the throughput of the algorithm  measured in (ms/frame) along with the energy consumed per frame (mJ/frame). EDP is the product of energy/frame and delay time. This way, a fair comparison can be made when deciding which hardware architecture is better suited for  specific computation. Lower EDP is better which means that the hardware architecture can finish specific computation tasks using less power in less time. 


\subsection{Measurement Techniques and Platforms:} \vspace{-1mm}

In this study, we evaluated two popular platforms for deploying embedded vision applications:
Nvidia Jetson TX2 and Xilinx ZCU102. These platforms come equipped with an on-board power measuring IC that can measure multiple power rails such as: CPU cores and GPU cores on the Jetson, and programmable logic, full power CPU cores and low power CPU cores on the FPGA platform.
On the Jetson TX2, shell scripts (running on its ARM CPU) sample power rails and log their values along with the system's timestamp into text files. The act of measuring power consumes power, thus consequently affects the results. The presented data in this paper has been corrected for this.
On the ZCU102, the Xilinx system controller tool (running on a PC) communicates over UART with a separate microcontroller (MSP430) that is controlling and reporting power data (so no correction needed).

For every benchmark, we first processed 1000 frames on the CPU core of the platform and then 1000 frames on the hardware accelerated part of the platform. This can be seen in Figure (\ref{figureMeasurementExample}), where the first two vertical lines mark the first 1000 frames on the CPU and the following two lines mark the last 1000 frames on the hardware. We computed the average frame rate by measuring the time between vertical lines and divided it by 1000. All frames were gray-scale with 1080p resolution. The x-axis represents the number of power samples taken for each platform. Note that the ZCU102 has a different sampling rate than the TX2.

\begin{figure}[b]
\centering
\captionsetup{justification=centering}
\setcounter{figure}{0}
\includegraphics[width=1\columnwidth, height=4cm]{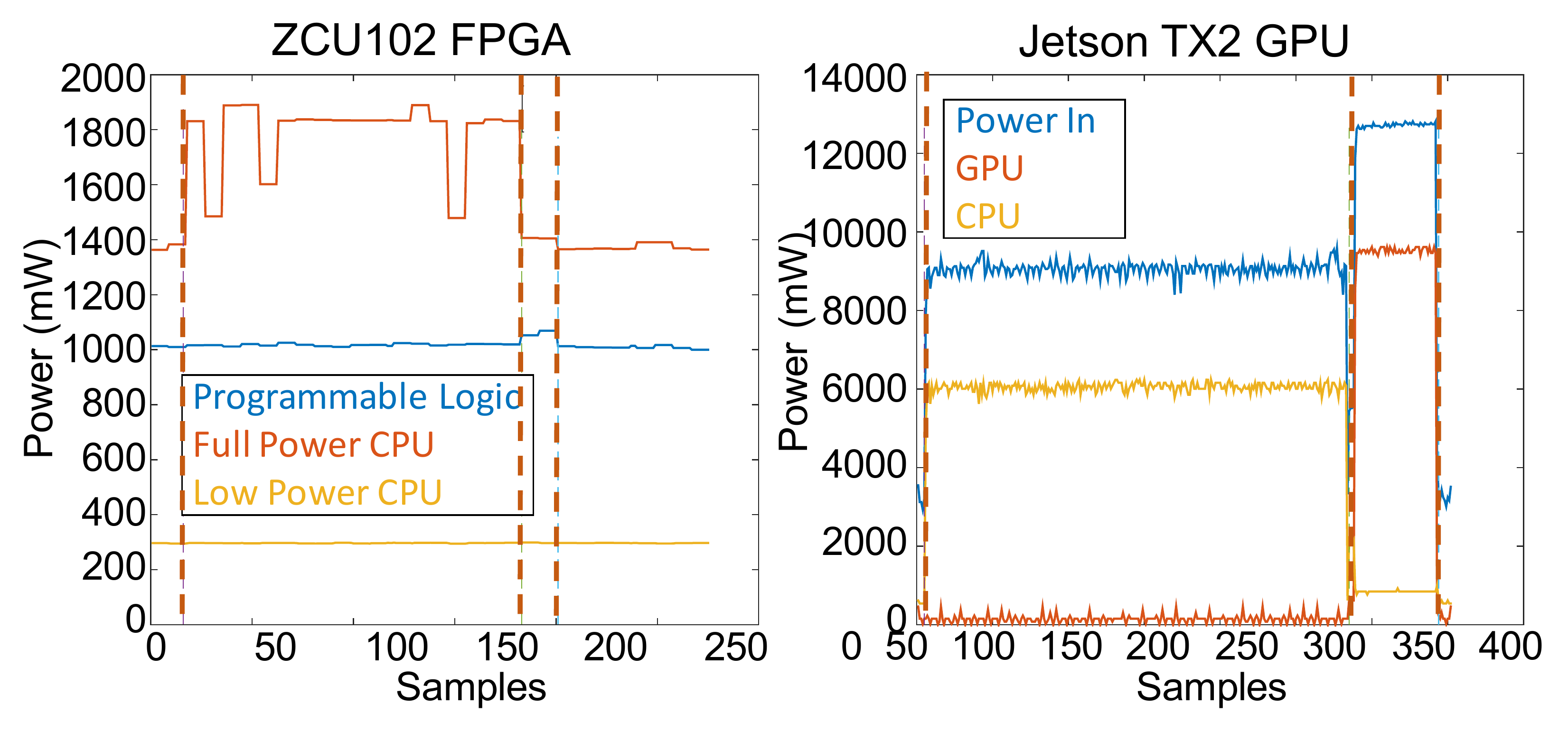}
\caption{ Measuring power samples on the platform's CPU cores (first 1000 frames), and its FPGA or GPU (second 1000 frames).}\vspace{-1em}
\label{figureMeasurementExample}
\end{figure}

\textbf{Hardware environments.} FPGA board: the Xilinx Zynq UltraScale+ MPSoC ZCU102 board  has a 16nm XCZU9EG FPGA, and  an on-board 4GB 64bit DDR4 RAM with a peak bandwidth of 136Gb/s.
GPU board: the Nvidia Jetson TX2 (Pascal 256 CUDA cores (16nm)) has 8GBs of 128bit DDR4 RAM with a peak bandwidth of 477.6 Gb/s. Both the FPGA and GPU have on-chip ARM CPU cores with NEON SIMD optimization.
The FPGA was clocked at 300 MHz, the ARM-A57 at 1.7 GHz, and the GPU at 998.4 MHz. 

\textbf{Software environments.}
 We used three publicly available vision libraries: (1) OpenCV 3.4 (2) Nvidia's VisionWorks 1.6 and (3) Xilinx's xfOpenCV 2018.3. While the OpenCV code base already comes with some GPU accelerated code, it does not come with FPGA support. For this purpose, we used OpenCV compatible C++ wrappers for xfOpenCV kernels \cite{PYNQCV}. With this wrapped functionality we were able to compile the same OpenCV code for both GPU and FGPA. Both OpenCV and VisionWorks support full IEEE floating-point precision, while xfOpenCV supports 8 bit precision.

\subsection{Benchmarking Approach} \vspace{-1mm}



In this study, we intentionally focused on evaluating the performance of out-of-the-box kernels from publicly available libraries (without writing special platform specific code around kernel calls) to give a fair comparison in terms of development efforts. For this reason, we first ran single kernel calls from OpenCV and VisionWorks libraries on the CPU and GPU, respectively, and instantiated a single kernel from xfOpenCV in FPGA fabric (even though small kernels utilize few FPGA resources). We then  measured the efficiency of representative vision pipelines on the three HW accelerators to quantify their speed and energy efficiency on these more complete vision applications.

For single kernel evaluation, we compared the efficiency of the HW accelerators in terms of their energy consumption per frame. We measured a vision kernel's dynamic power while excluding the static power required to power the rest of the platform. This better reflects the actual workload that is being deployed to the system since certainly for small kernels, the \textit{compute energy} \cite{giefers2016analyzing} (energy consumed for computation only) and data transfer energy are usually dominated by the static power.
In the vision pipeline evaluation, we compared the performance of HW accelerators in terms of their energy delay products (EDP). We used the total power consumption (static + dynamic), because it represents the actual power consumption when a complete system is deployed.
We also measured the maximum frame rate achieved on the three HW accelerators. The theoretical frame rate on the FPGA is fixed  for kernels that perform a single pass over the input image. Equation (\ref{equationFPGAfps}) shows an FPGA's frame rate when it is clocked at 300MHz for 1080p images. \vspace{-1em}

\begin{equation} \label{equationFPGAfps}
   FPS = \frac{300MHz}{1080\times1920\times 1pixel/cycle} = 144
\end{equation}

In order to have a sense of the amount of energy consumed for computation only, we measured the energy consumption of data movers in the FPGA and GPU. We implemented passthrough kernels which copy  an image's pixels from one memory location to another without applying any arithmetic/logical operations. In the FPGA implementation, Xilinx's SDx tool instantiates data movers \cite{boppana2015ultrascale} for each input or output port to transfer data between the memory mapped domain and the stream domain. Table \ref{tableDateMoverEnergyConsumption} shows that FPGA takes 6.945 ms to copy an entire image (1080p) with 0.41 mJ/frame, while GPU takes 1.298 ms with 0.19 mJ/frame. These values can be used to give a sense of the  ratio of energy consumed for computation to data transfer in each kernel.
\begin{center}
 \captionof{table}{Data Movers Energy Consumption Measurements}  
    \begin{tabular}{| L{1.8cm}|  C{2.6cm}| C{3cm} |  } \hline
   Platform & Time/frame (ms)  &  Energy/frame (mJ/f)   \\ \hline
   FPGA & 6.945   & 0.41  \\ \hline
   GPU  & 1.298   & 0.19   \\ \hline

    \end{tabular}
\label{tableDateMoverEnergyConsumption}
\end{center} \vspace{0.5em}

\section{Experimental Results} \vspace{-1mm} \label{sectionResults}

This section first presents the benchmarking results of single kernels from the six categories discussed in Section III. Then, a set of representative vision pipelines are evaluated.  \vspace{-1mm}

\subsection{Single Kernel Performance:} \vspace{-1mm}

Before evaluating the run-time performance and energy/frame consumption of single kernels on the HW accelerators, we first compare  two available GPU implementations: OpenCV CUDA module and Nvidia's VisionWorks toolkit. The OpenCV GPU module is written using CUDA and as a result benefits from the CUDA ecosystem. The Visionworks library applies many optimization techniques to boost performance, such as buffer reuse, kernel fusion, efficient use of streaming and CUDA textures, automatic scheduling across processing units, tiling and pipelining vision functions at the sub-frame level. Figure (\ref{figureCvCudaVisionWorks}) shows  the frame rate (bottom) and energy per frame (top) achieved by running vision kernels on the Jetson TX2. The Dark color represents OpenCV CUDA module, and the light color represents VisionWorks. We can observe that the VisionWorks implementation outperforms the OpenCV module in frame rate over all kernels. It achieved up to a 9.7$\times$ speedup compared to the OpenCV module. It also consumes less energy per frame over all kernels. It achieved up to a 6.3$\times$ reduction in energy consumption per frame. For this reason, in the rest of the paper, we will use only the VisionWorks implementation for the GPU.

\begin{figure*}[t]
\centering
\captionsetup{justification=centering}
\setcounter{figure}{1}
\begin{tikzpicture}
\begin{groupplot}
  [group style={group size= 1 by 2, xticklabels at=edge bottom}, height=3.5cm,width=18.5cm, ybar=5pt,
  x tick label style={rotate=35,anchor=east,font=\scriptsize},
  y tick label style={font=\small},
  xticklabels={split, combine, color conv, depth conv, threshold, absDiff, add/sub, and/or/xor, multiply, accumulate,squared, weighted, magnitude, phase, dilate, erode, boxFilter, filter2D, median, pyrDown, pyrUp, lookup, histogram, hist equl, integral, mean/std, min/max, resize, remap,affine warp, persp warp,canny, fast, harris, optical, stereoBM*},
  xtick={1,2,3,4,5,6,7,8,9,10,11,12,13,14,15,16,17,18,19,20,21,22,23,24,25,26,27,28,29,30,31,32,33,34,35,36}, ymajorgrids,    enlarge x limits=0.02
  ]
    \nextgroupplot[ylabel={Energy/F (mJ/f)}, bar width=4pt,
    ymin=0, ymax=8,
    restrict y to domain*=0:8.5, 
    nodes near coords={\pgfmathprintnumber{\rawy}},
    nodes near coords,every node near coord/.append style={font=\tiny,rotate=90,yshift=-0.2cm,xshift=0.3cm},
    nodes near coords align={vertical},
    visualization depends on=rawy\as\rawy, 
    nodes near coords={\pgfmathprintnumber{\rawy}},
    clip=false
    ]
    \addplot +[bar shift=-.0cm, BLACK, fill=Green3, area legend] coordinates {(1,4.38)(2,4.56)(3,4.69)(4,0.81)
    (5,0.87)(6,0.76)(7,0.64)(8,0.47)(9,1.09)(10,1.79)(11,3.37)(12,4.12)(13,2.27)(14,4.07)
    (15,73.7)(16,73.3)(17,4.02)(18,35.5)(19,24.4)(20,4.24)(21,20.6)    (22,2.57)(23,2.54)(24,1.96)(25,7.72)(26,2.1)(27,1.8)
    (28,0.6)(29,4.4)(30,4.3)(31,4.9)
    (32,35.4)(33,7.19)(34,45.6)(35,124.6)(36,113.1)};
    \addplot  +[bar shift=.15cm, BLACK, fill=Green1,postaction={pattern=horizontal lines, pattern color=Green2}]coordinates {(1,1.11)(2,1.01)(3,2.22)(4,0.57)
    (5,0.79)(6,0.72)(7,0.62)(8,0.54)(9,0.84)(10,0.91)(11,0.87)(12,0.91)(13,1.20)(14,1.27)
    (15,1.39)(16,1.39)(17,1.46)(18,2.92)(19,5.17)(20,1.35)(21,14.7)
    (22,0.72)(23,0.74)(24,1.71)(25,5.61)(26,1.92)(27,1.98)
    (28,0.28)(29,2.23)(30,1.91)(31,1.3)
    (32,28.7)(33,7.67)(34,7.18)(35,48.4)(36,0)};

    \nextgroupplot[ylabel={FPS}, bar width=4pt,
    legend style={at={(0.5,-0.55)},  anchor=north,legend columns=-1}, legend entries={CV::CUDA, VisionWorks},
    yshift=0.9cm,
    ymin=0, ymax=5900,
    nodes near coords, every node near coord/.append style={font=\tiny, rotate=90, yshift=-0.2cm, xshift=0.3cm},
    nodes near coords align={vertical} ]
    \addplot +[bar shift=.0cm, BLACK, fill=Green3, area legend] coordinates {(1,1049)(2,976)(3,522)(4,3289) (5,3279)(6,4082)(7,3861)(8,3745)(9,2967)(10,1253)(11,712)(12,491)(13,1225)(14,911)
    (15,63)(16,63)(17,932)(18,131)(19,110)(20,940)(21,212)    (22,1250)(23,1085)(24,1368)(25,456)(26,825)(27,730)
    (28,4926)(29,764)(30,962)(31,862)
    (32,84)(33,448)(34,104)(35,32)(36,39)};
    \addplot  +[bar shift=.15cm, BLACK, fill=Green1,postaction={pattern=horizontal lines, pattern color=Green2}, area legend]coordinates {(1,2278)(2,2132)(3,1170)(4,3344)
    (5,3745)(6,3717)(7,4065)(8,3546)(9,2985)(10,2262)(11,2283)(12,2174)(13,1838)(14,2222)
    (15,2639)(16,2646)(17,2825)(18,1285)(19,858)(20,2151)(21,2903)    (22,3058)(23,2165)(24,1350)(25,682)(26,1277)(27,734)
    (28,2128)(29,996)(30,2740)(31,2604)
    (32,101)(33,340)(34,502)(35,64)(36,0)  };
\end{groupplot}
\end{tikzpicture}
\caption{VisionWorks outperforms OpenCV CUDA module in terms of frame rate and energy/frame. *VisionWorks's implementation of the stereoBM kernel is not publicly available.} \vspace{-1em}
\label{figureCvCudaVisionWorks}
\end{figure*}
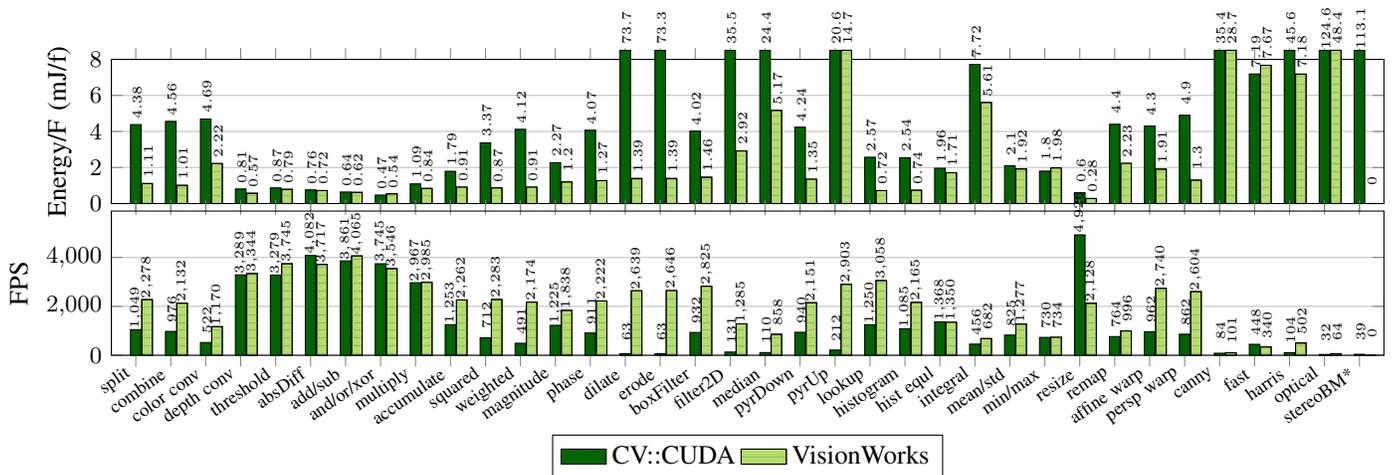

\begin{figure}[b]\vspace{-2em}
\setcounter{figure}{2}
\begin{tikzpicture}
\begin{groupplot}
  [group style={group size= 1 by 1, xticklabels at=edge bottom}, height=3.9cm, width=9cm, ybar=1pt, x tick label style={rotate=20,anchor=east},
  xticklabels={split, combine, color conv, depth conv},
  xtick={1,2,3,4}, ymajorgrids]

    \nextgroupplot[ylabel={Energy/Frame (mJ/f)}, bar width=5pt, nodes near coords,every node near coord/.append style={font=\tiny,rotate=90, yshift=-0.2cm,xshift=0.3cm},
    nodes near coords align={vertical},
    legend style={at={(0.5,-0.35)}, anchor=north,legend columns=-1},
    legend entries={ARM57,GPU,FPGA},
    ymin=0, ymax=5, restrict y to domain*=0:5, 
    ]
    \addplot +[bar shift=-.18cm, BLACK, fill=Blue1, area legend] coordinates {(1,3.1)(2,2.9)(3,2.4)(4,4.5)};
    \addplot  +[bar shift=-.0cm, BLACK, area legend, fill=Green1,postaction={pattern=horizontal lines, pattern color=Green2},  area legend
    ] coordinates {(1,1.14)(2,1.06)(3,1.1)(4,0.5)};
    \addplot  +[bar shift=.18cm, BLACK, fill=Red1, postaction={pattern=north east lines, pattern color=Red2}, area legend]coordinates {(1,1.2)(2,1.1)(3,1.3)(4,1.2)};
\end{groupplot}
\end{tikzpicture}
\caption{Input Processing Operations Kernels }  \vspace{-2em}
\label{figureInputProcessingOperationsKernels}
\end{figure}
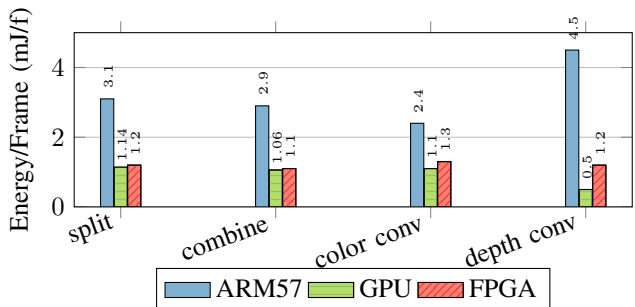

Next, we measured the energy per frame consumption of vision kernels from the following six categories: (1) input processing, (2) arithmetic operations, (3) filter operations, (4) image analysis,  (5) geometric transformation, and (6) composite kernels.

\textit{Input processing:} The energy/frame  of input processing kernels is shown in Figure (\ref{figureInputProcessingOperationsKernels}). These kernels mapped well to the GPU and FPGA compared to the CPU because of their significant  data parallelism, low complexity, and no data dependency. The GPU and FPGA achieved an average reduction ratio of 1.79$\times$ and 1.41$\times$ in energy/frame compared to the CPU. It also shows that GPU's implementation of bit-depth conversion achieved a 2.4$\times$ reduction compared to FPGA, because of the efficient use of streaming and CUDA textures in the VisionWorks kernel's implementation.

\begin{figure}[b]\vspace{-2em}
\begin{tikzpicture}
\begin{groupplot}
  [group style={group size= 1 by 1, xticklabels at=edge bottom}, height=3.9cm,width=9cm, ybar=0.6pt,
  x tick label style={rotate=21,anchor=east},
  xticklabels={threshold, absDiff, add/sub, and/or/xor, multiply, accumulate,squared, weighted, magnitude, phase},
  xtick={1,2,3,4,5,6,7,8,9,10},ymajorgrids
  ]
    \nextgroupplot[ylabel={Energy/Frame (mJ/f)}, bar width=5pt,
    nodes near coords,every node near coord/.append style={font=\tiny,rotate=90, yshift=-0.2cm,xshift=0.3cm},
    nodes near coords align={vertical}, legend style={at={(0.5,-0.35)},  anchor=north,legend columns=-1}, legend entries={ARM57,GPU,FPGA},
    ymin=0, ymax=4, restrict y to domain*=0:4.2, 
    visualization depends on=rawy\as\rawy, 
    nodes near coords={\pgfmathprintnumber{\rawy}},
    clip=false
    ]
    \addplot +[bar shift=-.19cm, BLACK, fill=Blue1, area legend] coordinates {(1,0.6)(2,1.0)(3,0.9)(4,0.9)(5,3.5)(6,2.7)(7,3.2)(8,3.4)(9,6.3)(10,6.1)};
    \addplot  +[bar shift=-.0cm, BLACK, fill=Green1,postaction={pattern=horizontal lines, pattern color=Green2}, area legend]coordinates {(1,0.8)(2,0.7)(3,0.6)(4,0.5)(5,0.8)(6,0.9)(7,0.8)(8,0.9)(9,1.2)(10,1.3)};
    \addplot  +[bar shift=.19cm, BLACK, fill=Red1,postaction={pattern=north east lines, pattern color=Red2}, area legend]coordinates {(1,0.8)(2,0.9)(3,0.8)(4,0.7)(5,1.3)(6,1.2)(7,1.1)(8,1.1)(9,1.3)(10,1.4)};

\end{groupplot}
\end{tikzpicture}
\caption{Arithmetic Operations Kernels}  \vspace{-2em}
\label{figureArithmeticOperationsKernels}
\end{figure}
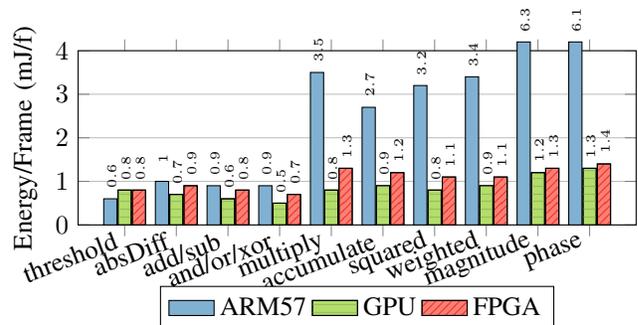

\textit{Image Arithmetic:} The performance of arithmetic/logic operations is shown in Figure (\ref{figureArithmeticOperationsKernels}). It shows that simple operations such as: threshold, absDiff, add/sub, and bitwise and/or/xor can be efficiently implemented by  the CPU. However, the CPU starts to perform poorly in kernels with multiplication operations, such as: multiply, accumulate squared, weighted, magnitude and phase.
The GPU has the lowest energy/frame compared to the CPU and FPGA. The GPU's implementations achieved an average reduction ratio in energy/frame of 4.6$\times$ and 7.2$\times$ compared to CPU and FPGA, respectively. An expected result, as  these algorithms can be granulated into many pieces that execute the same operation (SIMT).

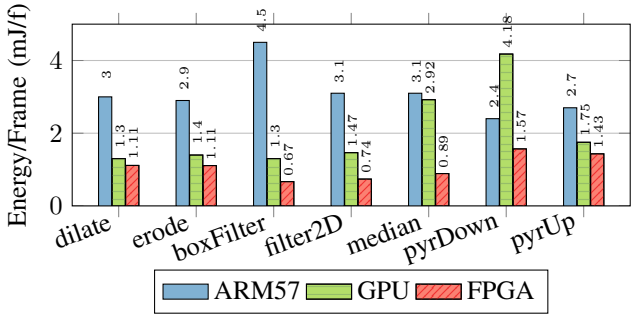
\begin{figure}[h]
\begin{tikzpicture}
\begin{groupplot}
  [group style={group size= 1 by 1, xticklabels at=edge bottom}, height=4cm,width=9cm, ybar=0.6pt,
  x tick label style={rotate=20,anchor=east},
  xticklabels={dilate, erode, boxFilter, filter2D, median, pyrDown, pyrUp},
  xtick={1,2,3,4,5,6,7},ymajorgrids]
    \nextgroupplot[ylabel={Energy/Frame (mJ/f)}, bar width=5pt,
    nodes near coords,every node near coord/.append style={font=\tiny,rotate=90, yshift=-0.2cm,xshift=0.3cm},
    nodes near coords align={vertical}, legend style={at={(0.5,-0.35)},  anchor=north,legend columns=-1}, legend entries={ARM57,GPU,FPGA},
    ymin=0, ymax=5, restrict y to domain*=0:5, 
    ]
    \addplot +[bar shift=-.18cm, BLACK, fill=Blue1, area legend] coordinates {(1,3.0)(2,2.9)(3,4.5)(4,3.1)(5,3.1)(6,2.4)(7,2.7)};
    \addplot  +[bar shift=-.0cm, BLACK, fill=Green1,postaction={pattern=horizontal lines, pattern color=Green2}, area legend]coordinates {(1,1.3)(2,1.4)(3,1.3)(4,1.465)(5,2.922)(6,4.178)(7,1.751)};
    \addplot  +[bar shift=.18cm, BLACK, fill=Red1,postaction={pattern=north east lines, pattern color=Red2}, area legend]coordinates {(1,1.114)(2,1.110)(3,0.668)(4,0.739)(5,0.890)(6,1.569)(7,1.433)};

\end{groupplot}
\end{tikzpicture}
\caption{Filters Operations Kernels } \vspace{-2em}
\label{figureFiltersOperationsKernels}
\end{figure}

\begin{figure}[h]
\begin{tikzpicture}
\begin{groupplot}
  [group style={group size= 1 by 1, xticklabels at=edge bottom}, height=4cm,width=9cm, ybar=0.6pt,
  x tick label style={rotate=20,anchor=east},
   xticklabels={lookup, histogram, hist equl, integral, mean/std, min/max}, xtick={1,2,3,4,5,6},ymajorgrids]
    \nextgroupplot[ylabel={Energy/Frame (mJ/f)}, bar width=5pt,
    nodes near coords,every node near coord/.append style={font=\tiny,rotate=90, yshift=-0.2cm,xshift=0.3cm},
    nodes near coords align={vertical}, legend style={at={(0.5,-0.35)},  anchor=north,legend columns=-1}, legend entries={ARM57,GPU,FPGA},
    ymin=0, ymax=6, restrict y to domain*=0:6.2, 
    visualization depends on=rawy\as\rawy, 
    nodes near coords={\pgfmathprintnumber{\rawy}},
    clip=false
    ]
    \addplot +[bar shift=-.18cm, BLACK, fill=Blue1, area legend] coordinates {(1,2.2)(2,4.3)(3,5.7)(4,4.5)(5,4.9)(6,6.5)};
    \addplot  +[bar shift=-.0cm, BLACK, fill=Green1,postaction={pattern=horizontal lines, pattern color=Green2}, area legend]coordinates {(1,0.7)(2,0.7)(3,1.7)(4,2.4)(5,2.9)(6,5.6)};
    \addplot  +[bar shift=.18cm, BLACK, fill=Red1,postaction={pattern=north east lines, pattern color=Red2}, area legend]coordinates {(1,0.6)(2,1.1)(3,1.1)(4,0.7)(5,0.8)(6,1.5)};

\end{groupplot}
\end{tikzpicture}
\caption{Image Analysis Operations Kernels }  \vspace{-1em}
\label{figureImageAnalysisOperationsKernels}
\end{figure}
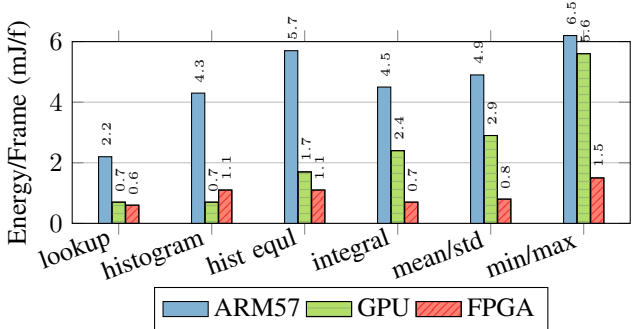

\textit{Image Filters:} In Figure (\ref{figureFiltersOperationsKernels}),  the results of filtering operations show that the FPGA performs better than the GPU and CPU for these kernels. The FPGA's implementation achieved an average reduction ratio of 1.8$\times$ and 7.4$\times$ in energy/frame compared to the GPU and CPU, respectively.
The memory access patterns and mathematical complexity of linear filters (filter2D, box filter, pyramid up and pyramid down) maps well to the parallel processing of the GPU and FPGA. Median filters, however, are unlike linear filters. They do not use sequential data access and multiply-and-accumulate operations, but sort  input elements and select the median of them, which makes them less straightforward to implement efficiently on a GPU. The morphological operations (dilate and erode) use  hit and miss functions over a structuring element. These functions are more difficult to implement than filtering functions due to comparison and branching. This explains the low frame rate  (as shown in Figure 2) and high energy/frame consumption of VisionWorks's implementations of  small (3$\times$3) filter kernels.

\textit{Image Analysis:}
The results of the image analysis kernels are shown in Figure (\ref{figureImageAnalysisOperationsKernels}).
For kernels such as lookup table, histogram, and histogram equalization, the energy/frame consumption of the FPGA achieves an average reduction of 1.2$\times$ compared to the GPU. While for kernels with more branching conditions and complex memory access patterns, such as integral image, mean/std, and min/max locations, the FPGA's implementation achieved an average reduction ratio of 3.5$\times$  compared to the GPU.

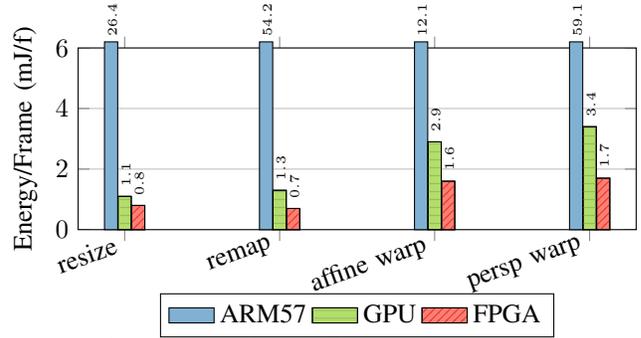
\begin{figure}[t]
\begin{tikzpicture}
\begin{groupplot}
  [group style={group size= 1 by 1, xticklabels at=edge bottom}, height=4cm,width=9cm, ybar=1pt,
  x tick label style={rotate=20,anchor=east},
  xticklabels={ resize, remap,affine warp, persp warp},
  xtick={1,2,3,4},ymajorgrids]
    \nextgroupplot[ylabel={Energy/Frame (mJ/f)}, bar width=5pt,
    nodes near coords,every node near coord/.append style={font=\tiny,rotate=90, yshift=-0.2cm,xshift=0.3cm},
    nodes near coords align={vertical}, legend style={at={(0.5,-0.35)},  anchor=north,legend columns=-1}, legend entries={ARM57,GPU,FPGA},
    ymin=0, ymax=6, restrict y to domain*=0:6.2, 
    visualization depends on=rawy\as\rawy, 
    nodes near coords={\pgfmathprintnumber{\rawy}},
    clip=false
    ]
    \addplot +[bar shift=-.18cm, black, fill=Blue1, area legend] coordinates {(1,26.4)(2,54.2)(3,12.1)(4,59.1)};
    \addplot  +[bar shift=-.0cm, black, fill=Green1,postaction={pattern=horizontal lines, pattern color=Green2}, area legend]coordinates {(1,1.1)(2,1.3)(3,2.9)(4,3.4)};
    \addplot  +[bar shift=.18cm, black, fill=Red1,postaction={pattern=north east lines, pattern color=Red2}, area legend]coordinates {(1,0.8)(2,0.7)(3,1.6)(4,1.7)};

\end{groupplot}
\end{tikzpicture}
\caption{Geometric Transforms Operations Kernels } \vspace{-2em}
\label{figureGeometricTransformsOperationsKernels}
\end{figure}

\textit{Geometric Transformation:}
The results of the geometric transformation kernels are shown in Figure (\ref{figureGeometricTransformsOperationsKernels}).
The CPU performs poorly for these kind of operations compared to the GPU and FPGA. Also, the FPGA was more energy efficient compared to the GPU. It achieved a reduction of 1.6$\times$ in energy/frame for the resize and remap kernels, and 2$\times$ for affine warp and perspective warp kernels. The computations in the warp operations are more complex compared to resize and remap as  mapping addresses need to be generated from 2$\times$3 or 3$\times$3 matrices before starting the mapping operation.

\begin{figure}[t]
\begin{tikzpicture}
\begin{groupplot}
  [group style={group size= 2 by 1, xticklabels at=edge bottom}, height=4cm,width=4.7cm, ybar=1pt,
  x tick label style={rotate=20,anchor=east},
  xtick={1,2,3,4,5},ymajorgrids]
    \nextgroupplot[ylabel={Energy/Frame (mJ/f)}, bar width=5pt,
    nodes near coords,every node near coord/.append style={font=\tiny,rotate=90, yshift=-0.2cm,xshift=0.3cm},
    nodes near coords align={vertical}, legend style={at={(1.0,-0.35)},  anchor=north,legend columns=-1}, legend entries={ARM57,GPU,FPGA},
    ymin=0, ymax=50, restrict y to domain*=0:50, 
    visualization depends on=rawy\as\rawy, 
    nodes near coords={\pgfmathprintnumber{\rawy}},
    clip=false, xticklabels={canny, fast, harris}
    ]
    \addplot +[bar shift=-.18cm, black, fill=Blue1, area legend] coordinates {(1,23.1)(2,22.9)(3,28.4)};
    \addplot  +[bar shift=-.0cm, black, fill=Green1,postaction={pattern=horizontal lines, pattern color=Green2}, area legend]coordinates {(1,18.7)(2,7.6)(3,7.1)};
    \addplot  +[bar shift=.18cm, black, fill=Red1,postaction={pattern=north east lines, pattern color=Red2}, area legend]coordinates {(1,4.9)(2,2.6)(3,1.9)};

    \nextgroupplot[ bar width=5pt,
    nodes near coords,every node near coord/.append style={font=\tiny,rotate=90, yshift=-0.2cm,xshift=0.3cm},
    nodes near coords align={vertical},
    ymin=0, ymax=210, restrict y to domain*=0:210, 
    visualization depends on=rawy\as\rawy, 
    nodes near coords={\pgfmathprintnumber{\rawy}},
    clip=false, xticklabels={optical, stereoBM}
    ]
    \addplot +[bar shift=-.18cm, black, fill=Blue1, area legend] coordinates {(1,197)(2,227)};
    \addplot  +[bar shift=-.0cm, black, fill=Green1,postaction={pattern=horizontal lines, pattern color=Green2}, area legend]coordinates {(1,44)(2,113)};
    \addplot  +[bar shift=.18cm, black, fill=Red1,postaction={pattern=north east lines, pattern color=Red2}, area legend]coordinates {(1,69)(2,30.8)};
\end{groupplot}
\end{tikzpicture}
\caption{Image Features, Optical Flow and Depth Kernels }  \vspace{-2em}
\label{figureImageFeaturesOpticalFlowDepthKernels}
\end{figure}
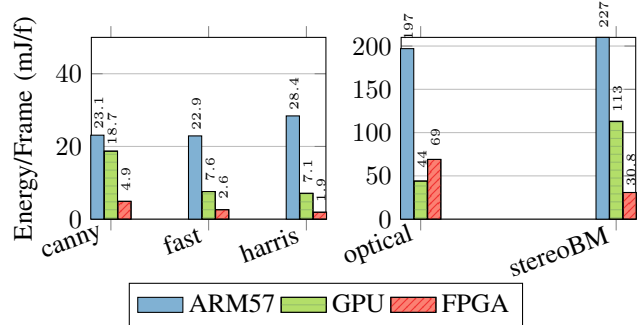

\textit{Composite Kernels:}
The last category in our study includes kernels for: (1) detecting image features (canny, fast and harris), (2) computing optical flow, and (3) computing disparity using  stereo block matching. Figure (\ref{figureImageFeaturesOpticalFlowDepthKernels}) shows that  the FPGA implementation of feature extraction kernels (canny, fast and harris) were more energy-efficient compared to the CPU and GPU  by an average reduction of 7.7$\times$ and 3.5$\times$, respectively.
The steps to calculate  sparse optical flow using the pyramid Lucas-Kanade algorithm includes extracting feature points from one frame and tracking them in the next frame. The FPGA implementation was able to detect 488 Harris corners compared to 94 for VisionWorks for the same input frame and parameters. Also, it was able to keep track of these points in the next frame. This explains the high energy/frame consumption in the FPGA implementation. Moreover, the VisionWorks's implementations of StereoBM is not open sourced yet, so the number reported in this paper is for the GPU implementation using OpenCV's CUDA module instead.

The average energy/frame reduction for the GPU and FPGA is shown in Table \ref{tableEnergyPerFrameReduction}. The ratio is with respect to CPU consumption (higher is better). We can observe a trend from simple kernels (top) to more complex kernels (bottom). The trend demonstrates that the performance of the GPU and FPGA compared to the CPU improves as kernels' complexity increases. For simple kernels (input processing and image arithmetic), the GPU shows the highest performance/energy efficiency, while for more complicated kernels (image filters, image analysis and geometric transform), the FPGA shows the highest performance/energy efficiency. Moreover, as the complexity of kernels increase, the FPGA shows higher energy-efficiency  compared to the GPU and CPU. This occurs due to the fact that more complex algorithms naturally occupy more resources on the programmable logic, as well as the fact that GPUs do not scale well for problems that are not easily divisible (data locality) or have many conditions or complex memory access patterns.

\begin{center} \small
 \captionof{table}{Ratios of Energy/Frame Reduction  (Reference CPU)}  \vspace{1mm}
    \begin{tabular}{| L{3.7cm}| C{1cm}| C{1cm} | C{1cm}|} \hline
   & CPU   &  GPU   & FPGA    \\ \hline
   Input Processing & 1 & \textbf{1.79$\times$}& 1.41$\times$  \\ \hline
   Image Arithmetic& 1 & \textbf{3.19$\times$} & 2.93$\times$  \\ \hline
   Image Filters& 1 & 3.17$\times$ & \textbf{3.89$\times$}  \\ \hline
   Image Analysis& 1 & 2.34$\times$ & \textbf{5.67$\times$}  \\ \hline
   Geometric Transform & 1 & 10.3$\times$ & \textbf{16.6$\times$}  \\ \hline
   Features/ OF/ StereoBM& 1 &7.44$\times$ & \textbf{22.3$\times$}  \\ \hline
    \end{tabular}
\label{tableEnergyPerFrameReduction}
\end{center}

\begin{figure*}[t]
\setcounter{figure}{8}
\centering
\captionsetup{justification=centering}
\begin{tikzpicture}
\begin{groupplot}
  [group style={group size= 1 by 1, xticklabels at=edge bottom}, height=4cm,width=18.5cm, ybar=5pt,
  x tick label style={rotate=35,anchor=east,font=\scriptsize},
  y tick label style={font=\small},
  xticklabels={split, combine, color conv, depth conv, threshold, absDiff, add/sub, and/or/xor, multiply, accumulate,squared, weighted, magnitude, phase, dilate, erode, boxFilter, filter2D, median, pyrDown, pyrUp, lookup, histogram, hist equl, integral, mean/std, min/max, resize, remap,affine warp, persp warp,canny, fast, harris, optical, stereoBM*},
  xtick={1,2,3,4,5,6,7,8,9,10,11,12,13,14,15,16,17,18,19,20,21,22,23,24,25,26,27,28,29,30,31,32,33,34,35,36}, ymajorgrids,    enlarge x limits=0.02]

    \nextgroupplot[ylabel={Frame/Second}, bar width=4pt,
    legend style={at={(0.5,-0.35)},  anchor=north,legend columns=-1}, legend entries={OpenCV(ARM57), VisionWorks(GPU)},
    yshift=0.9cm,
    ymin=0, ymax=4500,
    nodes near coords, every node near coord/.append style={font=\tiny, rotate=90, yshift=-0.2cm, xshift=0.3cm},
    nodes near coords align={vertical} ]
    \addplot +[bar shift=.0cm, BLACK, fill=Blue1, area legend] coordinates {(1,443)(2,881)(3,553)(4,1142) (5,1996)(6,904)(7,913)(8,880)(9,259)(10,301)(11,281)(12,287)(13,228)(14,150)

    (15,356)(16,360)(17,254)(18,322)(19,323)(20,378)(21,381)
    (22,962)(23,355)(24,309)(25,235)(26,336)(27,129)
    (28,69)(29,34)(30,70)(31,37)
    (32,65)(33,170)(34,11)(35,18)(36,12)};

    \addplot  +[bar shift=.15cm, BLACK, fill=Green1,postaction={pattern=horizontal lines, pattern color=Green2}, area legend]coordinates {(1,2278)(2,2132)(3,1170)(4,3344)
    (5,3745)(6,3717)(7,4065)(8,3546)(9,2985)(10,2262)(11,2283)(12,2174)(13,1838)(14,2222)
    (15,2639)(16,2646)(17,2825)(18,1285)(19,858)(20,2151)(21,2903)    (22,3058)(23,2165)(24,1350)(25,682)(26,1277)(27,734)
    (28,2128)(29,996)(30,2740)(31,2604)
    (32,101)(33,340)(34,502)(35,64)(36,39)  };
\end{groupplot}
\end{tikzpicture}
\caption{VisionWorks   achieved  an  average speed up over OpenCV of 2.9x, 4.2x, 6.3x, 3.9x, 42x and  4.5x in  the  six  categories of vision kernels. *OpenCV's CUDA implementation of stereoBM kernel is used (VisionWorks is not publicly available). } \vspace{-1em}
\label{figureCVVisionWorks}
\end{figure*}
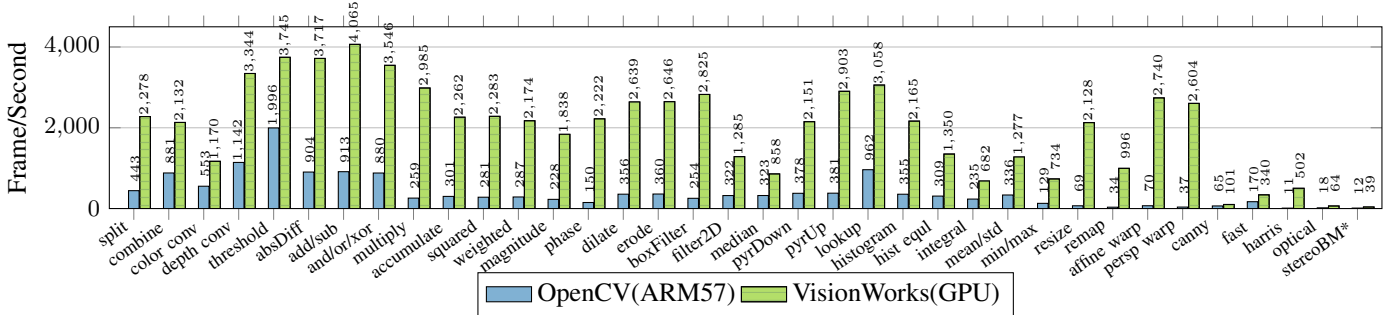

For completeness, Figure (\ref{figureCVVisionWorks}) includes a frame rate comparison between the ARM57 CPU OpenCV and GPU VisionWorks implementations. It shows that VisionWorks implementations achieved an average speed up over OpenCV implementation of  2.9$\times$, 4.2$\times$, 6.3$\times$, 3.9$\times$, 42$\times$ and 4.5$\times$ for the six categories of vision kernels: input processing, arithmetic operations, filter operations, image analysis, geometric transformation, image features, optical flow, and stereo block matching. The FPGA's frame rate met the theoretical rate of Equation (\ref{equationFPGAfps}) for kernels performing a single pass over the input image. The   theoretical frame rate for the FPGA is 144 fps when it is  clocked at 300MHz for 1080p resolution images.

\subsection{Complete Vision Pipeline Performance:}

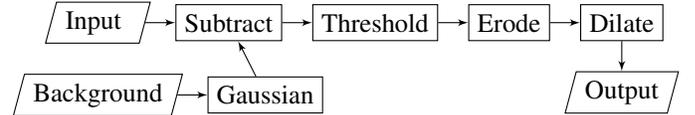
\begin{figure}[h]
\begin{tikzpicture}[node distance=4mm, >=latex',
        block/.style = {draw, rectangle, minimum height=4mm, minimum width= 10mm,align=center},
        tblock/.style = {draw, trapezium, minimum height=5mm,
                         trapezium left angle=75, trapezium right angle=105, align=center}]
            \node [tblock]                  (Input)     {Input };
            \node [block, right=of Input]   (Subtract)     {Subtract};
            \node [block, right=of Subtract]   (Thresh) {Threshold};
            \node [tblock, below=of Input]  (Model)      {Background};
            \node [block, right=of Model]  (Gaussian)      {Gaussian};
            \node [block, right=of Thresh]   (Erode)  {Erode};
            \node [block, right=of Erode]    (Dilate)  {Dilate};
            \node [tblock, below=of Dilate]  (Output)     {Output};

            \path[draw,->]  (Input)     edge (Subtract)
                            (Subtract)  edge (Thresh)
                            (Thresh)  edge (Erode)
                            (Erode)  edge (Dilate)
                            (Dilate)  edge (Output)
                            (Gaussian)  edge (Subtract)
                            (Model)  edge (Gaussian)  ;
\end{tikzpicture}
\caption{Background Subtraction Pipeline Components} 
\label{figureBackgroundSubtractionPipeline}
\end{figure}

In this section, we evaluated the performance of the HW accelerators for four representative pipelines. Common steps in many computer vision pipelines include: pre-processing, feature extraction, and post-processing. The pipelines used in our study follow this structure: (1) background subtraction, (2) color segmentation, (3) stereo block matching, and (4) Harris corner tracking. These pipelines are implemented on the GPU using VisionWorks OpenVX graph mode to enable its advanced optimization techniques (buffer reuse, kernel fusion, etc.). We also pipelined the execution of kernels on the FPGA at pixel/frame level using xfOpenCV modules. In this way, the FPGA can leverage the fact that image pixels stays within the programmable fabric and avoids going back and forth to read/write from external memory. The pipelines evaluated in this paper are:

\textbf{1. Background Subtraction:}
The background subtraction pipeline is used  to detect changes in image sequences \cite{Opencv}. It is mainly used when regions of interest  are foreground objects. The pipeline components include: subtraction, Gaussian filtering, threshold, erode and dilate, as shown in Figure (\ref{figureBackgroundSubtractionPipeline}).

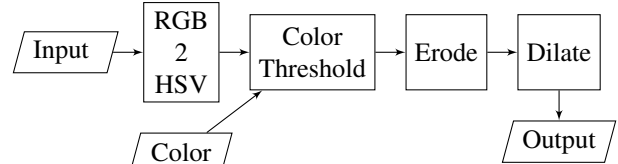
\begin{figure}[h]
\begin{tikzpicture}[node distance=4mm, >=latex',
    block/.style = {draw, rectangle, minimum height=10mm, minimum width= 10mm,align=center},
    tblock/.style = {draw, trapezium, minimum height=5mm,
                     trapezium left angle=75, trapezium right angle=105, align=center}]
        \node [tblock]                  (input)     {Input };
        \node [block, right=of input]   (RGB2HSV)   {RGB\\2\\HSV};
        \node [tblock, below=of RGB2HSV]  (color)     {Color};
        \node [block, right=of RGB2HSV]   (thresh)   {Color\\Threshold};
        \node [block, right=of thresh]   (Erode)  {Erode};
        \node [block, right=of Erode]     (Dilate)  {Dilate};
        \node [tblock, below=of Dilate]  (Output)     {Output};

        \path[draw,->]  (input)   edge (RGB2HSV)
                        (color)   edge (thresh)
                        (RGB2HSV) edge (thresh)
                        (thresh)  edge (Erode)
                        (Erode) edge (Dilate)
                        (Dilate) edge (Output);
\end{tikzpicture}
\caption{Color Segmentation Pipeline Components}\vspace{-2mm}
\label{figureColorSegmentationPipeline}
\end{figure}

\textbf{2. Color Segmentation:} This pipeline is used to partition  an image into multiple segments based on a specific range of colors. It converts the color format from RGB to HSV, then applies range thresholding to its three channels, and applies erode and dilate operations, as shown in Figure (\ref{figureColorSegmentationPipeline}).  \vspace{-1em}


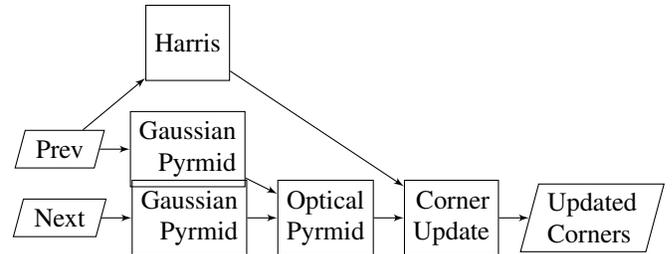
\begin{figure}[h]
\begin{tikzpicture}[node distance=4mm, >=latex',
    block/.style = {draw, rectangle, minimum height=10mm, minimum width= 10mm,align=right},
    tblock/.style = {draw, trapezium, minimum height=5mm,
                     trapezium left angle=75, trapezium right angle=105, align=left}]

        \node [tblock]                  (Prev)  {Prev};
        \node [tblock, below=of Prev]   (Next)  {Next};
        \node [block, right=of Prev]    (Gaussian1)  {Gaussian\\Pyrmid};
        \node [block, right=of Next]    (Gaussian2)  {Gaussian\\Pyrmid};
        \node [block, above=of Gaussian1]      (Harris)     {Harris};
        \node [block, right=of Gaussian2]   (Optical)  {Optical\\Pyrmid};
        \node [block, right=of Optical]   (Corner)  {Corner\\Update};
        \node [tblock, right=of Corner]   (Updated)  {Updated\\Corners};

        \path[draw,->]  (Prev)   edge (Gaussian1)
                        (Next)   edge (Gaussian2)
                        (Gaussian1)   edge (Optical)
                        (Gaussian2)   edge (Optical)
                        (Optical)   edge (Corner)
                        (Corner)   edge (Updated)
                        (Prev)   edge (Harris)
                        (Harris)   edge (Corner) ;
        \end{tikzpicture}
\caption{Harris Corners Tracking Pipeline Components} \vspace{-0.5em}
\label{figureHarrisCornersTrackingPipeline}
\end{figure}

\textbf{3. Harris Corners Tracking:}
This pipeline is used to detect and track feature points in a set of successive frames of a video. It takes in the current and next frame as  inputs. It computes Harris corners from the current frame and outputs a list of tracked corners in the next frame. The pipeline uses five kernels as shown in Figure (\ref{figureHarrisCornersTrackingPipeline}).   \vspace{-1em}


\begin{figure}[h]
\begin{tikzpicture}[node distance=4mm, >=latex',
    block/.style = {draw, rectangle, minimum height=10mm, minimum width= 10mm,align=center},
    tblock/.style = {draw, trapezium, minimum height=4.5mm,
                     trapezium left angle=75, trapezium right angle=105, align=center}]
        \node [tblock]                  (left)      {Left };
        \node [tblock, below=of left]   (right)     {Right};
        \node [block, right=of left]    (undistort1)  {Undistort\\Rectify};
        \node [block, right=of right]   (undistort2)  {Undistort\\Rectify};
        \node [block, right=of undistort1]   (remap1)  {Remap};
        \node [block, right=of undistort2]   (remap2)  {Remap};
        \node [block, right=of remap2]   (stereoBM)  {Stereo\\BM};
        \node [tblock, right=of stereoBM]   (disparity)  {disparity};

        \path[draw,->]  (left)   edge (undistort1)
                        (right)  edge (undistort2)
                        (undistort1)  edge (remap1)
                        (undistort2)  edge (remap2)
                        (remap1)  edge (stereoBM)
                        (remap2)  edge (stereoBM)
                        (stereoBM)  edge (disparity);
\end{tikzpicture}
\caption{Stereo Block Matching Pipeline Components} \vspace{-0.5mm}
\label{figureSteroBMPipeline}
\end{figure}
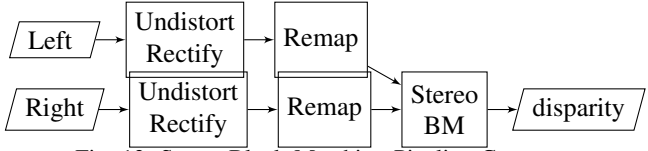

\newpage
\textbf{4. Stereo Block Matching:}
This pipeline is used to generate a disparity map given the camera parameters and inputs from a stereo camera setup. It is used as a first step in creating a three dimensional map of an environment. The main components involved in the pipeline are shown in Figure (\ref{figureSteroBMPipeline}). It consists of stereo rectification, remapping, and disparity estimation using a local block matching method.


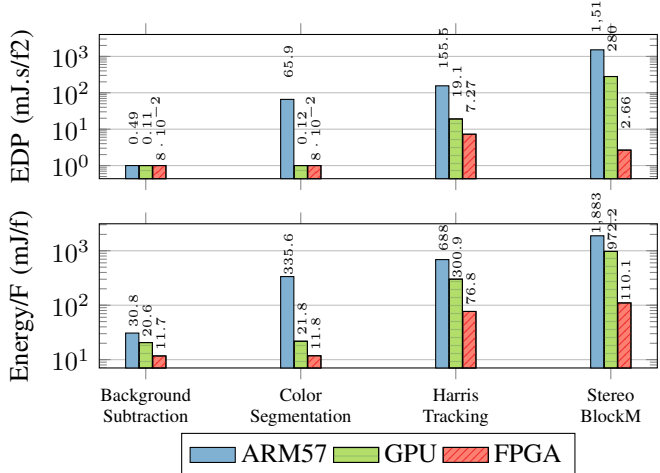
\begin{figure}[h]
\captionsetup{justification=centering}
\begin{tikzpicture}
\begin{groupplot}
  [group style={group size= 1 by 2, xticklabels at=edge bottom}, height=3.5cm,width=9cm, ybar=1pt,
 symbolic x coords={ Background Subtraction, Color Segmentation, Harris Tracking, Stereo BlockM}, xticklabel style={text width=1.5cm, font=\scriptsize, align=center}, xtick=data, ymajorgrids
    ,ymode=log, log basis y={10},log origin=infty
    ]
    \nextgroupplot[ylabel={EDP (mJ.s/f2)}, bar width=5pt,
    nodes near coords,every node near coord/.append style={font=\tiny,rotate=90, yshift=-0.2cm,xshift=0.5cm}, yshift=-0.9cm,
    nodes near coords align={vertical},
    ymin=0, ymax=4000, restrict y to domain*=0:4000, 
    visualization depends on=rawy\as\rawy, 
    nodes near coords={\pgfmathprintnumber{\rawy}},
    clip=false]
    \addplot +[bar shift=-.18cm, black, fill=Blue1, area legend] coordinates
    {(Background Subtraction,0.49)(Color Segmentation,65.9)(Harris Tracking,155.5)(Stereo BlockM,1516)};
    \addplot  +[bar shift=-.0cm, black, fill=Green1,postaction={pattern=horizontal lines, pattern color=Green2}, area legend]coordinates {(Background Subtraction,0.11)(Color Segmentation,0.12)(Harris Tracking,19.1)(Stereo BlockM,280)};
    \addplot  +[bar shift=.18cm, black, fill=Red1,postaction={pattern=north east lines, pattern color=Red2}, area legend]coordinates {(Background Subtraction,0.08)(Color Segmentation,0.08)(Harris Tracking,7.27)(Stereo BlockM,2.66)};

    \nextgroupplot[ylabel={Energy/F (mJ/f)}, bar width=5pt,
    nodes near coords,every node near coord/.append style={font=\tiny,rotate=90, yshift=-0.2cm,xshift=0.3cm},
    yshift=0.4cm,
    nodes near coords align={vertical},
    legend style={at={(0.5,-0.45)},  anchor=north,legend columns=-1}, legend entries={ARM57,GPU,FPGA},
    visualization depends on=rawy\as\rawy, 
    nodes near coords={\pgfmathprintnumber{\rawy}},
    clip=false]

    \addplot +[bar shift=-.18cm, black, fill=Blue1, area legend] coordinates
    {(Background Subtraction,30.8)(Color Segmentation,335.6)(Harris Tracking,688)(Stereo BlockM,1883)};
    \addplot  +[bar shift=-.0cm, black, fill=Green1,postaction={pattern=horizontal lines, pattern color=Green2}, area legend]coordinates
    {(Background Subtraction,20.6)(Color Segmentation,21.8)(Harris Tracking,300.9)(Stereo BlockM,972.2)};
    \addplot  +[bar shift=.18cm, black, fill=Red1, postaction={pattern=north east lines, pattern color=Red2}, area legend]coordinates
    {(Background Subtraction,11.7)(Color Segmentation,11.8)(Harris Tracking,76.8)(Stereo BlockM,110.1)};
\end{groupplot}
\end{tikzpicture}
\caption{FPGA outperforms GPU and CPU in energy/frame consumption and EDP } \vspace{-1em}
\label{figureEnergyPerFramePipelines}
\end{figure}

Figure (\ref{figureEnergyPerFramePipelines}) plots the Energy/frame and EDP comparison of the four pipelines, and shows the FPGA implementations consume less energy/frame compared to the CPU and GPU  for all pipelines. The FPGA is also more efficient in terms of EDP (lower EDP is better). The FPGA's Energy/frame and EDP reduction ratio with respect to the GPU is listed in Table \ref{tableEnergyPerFrameReduction2}. As the complexity of the pipeline grows, the energy/frame and EDP reduction ratio increases. More complex vision pipelines can use more of the FPGA programmable logic, reducing the relative impact of static power consumption. Additionally, data communicated between modules of the pipeline are kept on-chip in the streaming FPGA implementation.\vspace{-2em}

\begin{center} \small
 \captionof{table}{FPGA's Reduction Ratios  with repsect to GPU}  
    \begin{tabular}{| L{3.2cm}|  C{2.2cm} | C{2cm}|} \hline
       Pipeline &   Energy/frame (mJ/f) & EDP\hspace{5mm}(mJ.s/f2)   \\ \hline
       Background Subtraction   & 1.74$\times$  & 1.32$\times$  \\ \hline
       Color Segmentation       & 1.86$\times$  & 1.41$\times$  \\ \hline
       Harris Corners Tracking  & 3.94$\times$  & 2.65$\times$   \\ \hline
       Stereo Block Matching    & 8.83$\times$  & 107.7$\times$   \\ \hline
        \end{tabular}
\label{tableEnergyPerFrameReduction2}
\end{center}

\section{Conclusion} \vspace{1mm} \label{sectionConclusions}

In this paper, we benchmarked algorithms from all the computer vision categories defined by the open standard, OpenVX, on both GPU- and FPGA-accelerated embedded platforms. We found that while many simple and easy-to-parallelize kernels perform well on GPUs (1.1--3.2$\times$ energy/frame reduction), for more complete vision pipelines, FPGAs outperform GPUs and CPUs (1.2--22.3$\times$ energy/frame reduction).  Moreover, FPGAs perform increasingly better as the complexity of vision pipelines grow.
This is evidenced by the energy-delay product, a metric that takes into account not only the energy/frame, but also algorithm throughput. 
Our future work will extend this analysis to the latest platform generation, like Nvidia's recently released AGX board, and will extend this benchmarking suite with key modules from machine learning and mixed pipelines composed of vision  and machine learning kernels.

\bibliographystyle{ieeetr}
\bibliography{References}
\end{document}